\pdfoutput=1

\documentclass[11pt]{article}

\usepackage{emnlp2021}
\usepackage{times}
\usepackage{latexsym}
\usepackage{graphicx}
\usepackage[T1]{fontenc}

\usepackage[utf8]{inputenc}
\usepackage[ruled]{algorithm2e}
\usepackage{algorithmic}
\usepackage{multirow}
\usepackage{multicol}
\usepackage{amsmath}
\usepackage{xcolor}

\usepackage{microtype}

\usepackage{ulem}

\usepackage{microtype}

\title{How to Select One Among All ? An Extensive Empirical Study Towards the Robustness of Knowledge Distillation in Natural Language Understanding}

\author{Tianda Li$^1$, Ahmad Rashid$^1$, Aref Jafari$^{1,2}$, Pranav Sharma$^2$,\\ \textbf{Ali Ghodsi$^3$, Mehdi Rezagholizadeh$^1$} \\
$^1$Huawei Noah’s Ark Lab\\
$^2$David R. Cheriton School of Computer Science, University of Waterloo\\
$^3$Department of Statistics and Actuarial Science, University of Waterloo \\

{\texttt \{tianda.li, ahmad.rashid, mehdi.rezagholizadeh\}@huawei.com}; { {aref.jafari@uwaterloo.ca}}
}

\begin{document}
\maketitle
\begin{abstract}

Knowledge Distillation (KD) is a model compression algorithm that helps transfer the knowledge of a large neural network into a smaller one. Even though KD has shown promise on a wide range of Natural Language Processing (NLP) applications, little is understood about how one KD algorithm compares to another and whether these approaches can be complimentary to each other.
In this work, we evaluate various KD algorithms on in-domain, out-of-domain and adversarial testing. We propose a framework to assess the adversarial robustness of multiple KD algorithms. Moreover, we introduce a new KD algorithm, Combined-KD~\footnote{We will release our code at https://github.com/huawei-noah/KD-NLP.}, which takes advantage of two promising approaches (better training scheme and more efficient data augmentation). Our extensive experimental results show that Combined-KD achieves  state-of-the-art results on the GLUE benchmark, out-of-domain generalization, and adversarial robustness compared to competitive methods.

\end{abstract}


\section{Introduction}
Pre-trained language models have achieved impressive results on a wide variety of NLP problems~\cite{Peters:2018,devlin2019bert,liu2019roberta,yang2020xlnet,Radford2018ImprovingLU,radford2019language}. 
 The rapidly increasing parameter size, however, has not only made the training process more challenging~\cite{ghaddar2019contextualized}, but has also become an obstacle when deploying these models on edge devices.
To address the over-parameterization and computation cost of state-of-the-art (SOTA) pre-trained language models,
KD~\cite{hinton2015distilling} has emerged as a widely used model compression technique in the
literature~\cite{rogers-etal-2020-primer}. 

Recent work on improving KD can be categorized into two directions: 1) Designing a better training scheme to help the student model learn efficiently from the teacher model. E.g., matching  the student model's intermediate weights and attention matrices with the teacher's during the training~\cite{sun2019patient, wang2020minilm, passban2020alpkd} or designing progressive or curriculum based learning~\cite{Aref2021,sun2020mobilebert,mirzadeh2020improved} to overcome capacity gap~\cite{mirzadeh2020improved} between teacher and student models.
2) Employing data-augmentation~\cite{jiao-etal-2020-tinybert,fu2020rolewise,rashid-etal-2021-mate,kamalloo-etal-2021-far} to improve KD by using more diverse training data. It is difficult to compare these methods since, typically, the teachers and students are initialized differently.  

The robustness of KD also requires further investigation. Recent studies have revealed that the strong performance of neural networks in NLP can be partially attributed to learning spurious statistical patterns in the training set and even the SOTA models can make mistakes if a few words in their input are replaced~\cite{jin2019bert,li2020bertattack,8424632,DBLP:journals/corr/abs-1904-12104}. As a result, even though KD has achieved good performance in different downstream tasks~\cite{jiao-etal-2020-tinybert, sanh2020distilbert, sun2020mobilebert}, it is desirable to investigate if these KD methods learn semantic knowledge and are robust enough to retain their performance on an out-of-domain (OOD) dataset~\cite{mccoy-etal-2019-right,zhang-etal-2019-paws} or under an adversarial attack~\cite{jin2019bert,8424632}. 
It would also be desirable to evaluate if different KD algorithms are complimentary and can be combined successfully.

Our contributions in this paper are as follows:

\begin{enumerate}

    \item We compare KD algorithms for BERT compression initialized with the same teacher and student, and rank them against one another on the GLUE benchmark.
    \item  We conduct OOD and adversarial evaluation to investigate the robustness of  KD methods.
    \item  We propose a unified adversarial framework (UAF) that can evaluate adversarial robustness in a multi-model setting to fairly compare different models.
    \item We introduce a new KD method named Combined-KD (ComKD) by taking advantage of data-augmentation and progressive training. Results show that our proposed ComKD not only achieves a new SOTA on the GLUE benchmark, but is also more robust compared to competitive KD baselines under OOD evaluation and adversarial attacks.
    

\end{enumerate}


\section{Related Work}
\subsection{Unsupervised pre-training}
Unsupervised pre-training~\cite{devlin2019bert,liu2019roberta,yang2020xlnet} has been shown to be very effective in improving the performances of a wide range of NLP problems. 
Model performance has scaled well with larger number of parameters and training data~\cite{raffel2020exploring,brown2020language,radford2019language}. 

\subsection{Knowledge distillation (KD)}
Knowledge distillation~\cite{hinton2015distilling,bucilua2006model,8424632,kamalloo-etal-2021-far,rashid2020towards,wu-etal-2020-skip} has emerged as an important algorithm in language model compression \cite{jiao-etal-2020-tinybert, sanh2020distilbert, sun2020mobilebert}. In the general setting, a larger model is employed as the teacher and a smaller model as the student, and the knowledge of the teacher is transferred to the student during the KD training. Specifically, in addition to a supervised
training loss, the student also considers a distillation loss over the soft target probabilities of the teacher.

~\citet{sun2019patient} proposed distilling intermediate layer representation in addition to the regular distillation loss. Since the teacher typically has more layers than the student, the algorithm has to decide which layers to distil from and which to skip.~\citet{passban2020alpkd} overcome this challenge by designing an attention mechanism which fuses teacher-side information and takes each layer’s significance into consideration. ~\citet{Aref2021} identifies the capacity gap problem \cite{mirzadeh2020improved} i.e., as the teacher increases in size (and performance), the performance of a fixed size student will initially improve and then drop down.
 They propose to improve KD by using temperature to anneal the teacher's output gradually, then the student will be trained following the annealed output. ~\citet{rashid-etal-2021-mate} proposed adversarial data augmentation to improve KD. They train a generator to perturb data samples so as to increase the divergence between the student and teacher output.

\subsection{Model Robustness Evaluation}

It has been demonstrated that models which are SOTA on different NLP applications, such as machine translation and natural language understanding, can be brittle to small perturbations of the data~\cite{cheng2019robust, belinkov2017synthetic, mccoy-etal-2019-right}. In our work we consider OOD tests and adversarial attacks. 

\subsubsection{Out-of-Domain test}
The purpose of the OOD test is to change the distribution of dataset by applying fixed patterns to the original dataset. E.g., \citet{mccoy-etal-2019-right} used three heuristic rules to modify the MNLI evaluation set. \citet{zhang-etal-2019-paws} proposed well-formed paraphrase and non-paraphrase pairs with high lexical overlap based on the original QQP~\cite{DBLP:journals/corr/abs-1804-07461} dataset. \citet{DBLP:journals/corr/abs-1805-02266} introduced a natural language inference (NLI) test set by replacing a single word of a training instance using WordNet~\cite{Miller1995WordNetAL}. 


\subsubsection{Adversarial Attack}
Adversarial examples, which were first identified in computer vision~\cite{Goodfellow2015ExplainingAH,DBLP:journals/corr/KurakinGB16,labacacastro2021universal}, are small perturbations to data which are indiscernible for humans but can confuse a neural network classifier. The standard approach is to add gradient-based perturbation on continuous input spaces~\cite{Goodfellow2015ExplainingAH,DBLP:journals/corr/KurakinGB16}. Recently, studies also explore the use of adversarial examples on NLP tasks, e.g., using adversarial examples to measure robustness against an adversarial attack~\cite{jin2019bert}, or adding adversarial examples during training process to help models improve in robustness and generalization~\cite{zhu2020freelb,ghaddar2021context,ghaddar-etal-2021-end,rashid-etal-2021-mate}.~\citet{jin2019bert} proposed a model dependent framework, textfooler, to generate adversarial samples to attack existing models. Different from previous rule-based frameworks, textfooler can automatically replace the most semantically important words based on a specific model's output.


\section{Methodology}
\begin{figure}[tb]
    \centering
    
    \includegraphics[width=8cm]{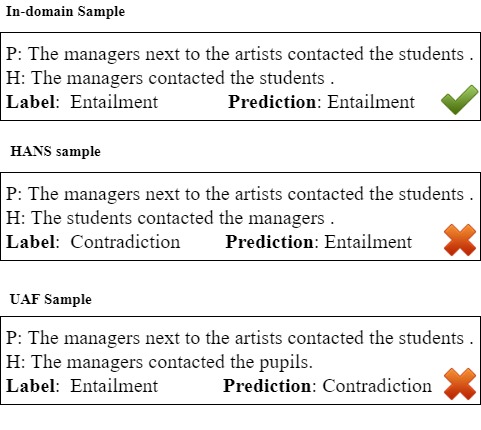}
    \caption{Different model evaluations for Natural Language Inference. The model is evaluated on in-domain, out-of-domain and adversarial samples and makes an error on the latter two. }
    \label{fig:SAMPLE_UAF}
    
\end{figure}
\label{sec:experiments}

First, we evaluate different KD algorithms on three kinds of test sets. In-domain testing where the training and test sets are from the same distribution, OOD test where the test set is specially designed to diagnose whether the model overfits to spurious lexical patterns and an adversarial test set to measure robustness to adversarial examples. Then we present our ComKD algorithm.

In figure~\ref{fig:SAMPLE_UAF}, we present an example from Natural Language Inference where the model can predict an in-domain sample correctly, but makes mistakes on OOD (HANS) and adversarial (UAF) evaluation. Specifically, in this example, the HANS sample changes the object and the subject whereas the UAF sample replaces the most semantically important word.


\subsection{In-domain test}
We train on the GLUE benchmark~\cite{DBLP:journals/corr/abs-1804-07461} and use the provided evaluation sets as our in-domain test datasets. We evaluate both on the GLUE dev set and the test set. 

\subsection{Out-of-Domain test}
We employ HANS~\cite{zhang-etal-2019-paws} and PAWS~\cite{DBLP:journals/corr/abs-1904-01130} as our OOD test set. Models are trained on MNLI and QQP datasets respectively.

\subsection{Adversarial Test}
\label{sec:uniontextfooler}

\begin{figure}[tb]
    \centering
    \includegraphics[width=7cm]{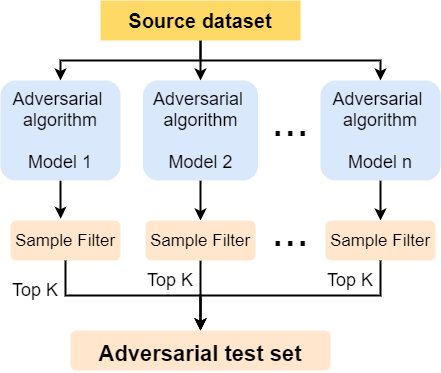}
    \caption{Unified Adversarial Framework}
    \label{fig:Adversarial}
\end{figure}

Adversarial attack is an effective way to test the robustness of a model. 
Current adversarial attacks, however, focus on single model attacks which can not be used to draw a comparison between different models directly.
To deal with this, we propose a
unified adversarial framework (UAF), presented in figure~\ref{fig:Adversarial}, which can help us fairly compare different KD algorithms with the same adversarial attack.

After selecting the adversarial algorithm and source dataset, each model that is used in the evaluation will apply the same adversarial algorithm to generate adversarial samples. To keep the quality of generated samples during the generation, a quality score will be employed to rank the adversarial samples.
The quality score will be computed follow the function below:
\begin{equation}
Score = cos(Model_{n}(X), Model_{n}(X')),
\end{equation}
Where $Model_{n}$ is the model that generates the adversarial sample, $X$ is the original sample, $X'$ is the generated adversarial sample. We calculate the cosine distance between the hidden state of the two \emph{[CLS]} tokens  to get the quality score. 
Intuitively, adversarial examples are not expected to be too similar to the original sample or the models can easily distinguish it. On the other hand, the adversarial example can not be too distant from the original sample, as it will compromise model quality.
As a result, for the sample filter step, two threshold values $\lambda_{up}$ and $\lambda_{down}$ will be used to filter the adversarial samples. Only the sample with a quality score in the range of $( \lambda_{down}, \lambda_{up}]$ will be kept.
Finally, we collect top K best samples from each sample filter to complete our adversarial test set.

\subsection{Combined Knowledge Distillation}

According to the current results in the literature, MATE-KD~\cite{rashid-etal-2021-mate} and Annealing-KD~\cite{Aref2021} are two of the best methods from the two family of KD algorithms. We will attempt to combine their strengths and evaluate whether it will improve the performance overall. 

The teacher logit annealing of Annealing-KD is specially interesting because it addresses the capacity gap problem. Pre-trained language models are constantly increasing in size and larger model tend to perform better on downstream tasks. As demonstrated by ~\citet{mirzadeh2020improved}, the performance of a fixed student does not necessarily scale with a better teacher. The adversarial algorithm in MATE-KD, on the other hand, augments data which is designed to probe parts of the teacher function not explored by the training data. Other advantages of these methods are that they only distil the teacher logits (as opposed to the weights and attention maps), do not introduce additional hyper-parameters, and perform well empirically.

We employ a masked language model (MLM) generator for data augmentation. The generator is trained to produce samples which maximize the divergence between the teacher and the student logits. Additionally, the generator fixes most of the text and only generates the masked tokens so that the text does not diverge too much from the training distribution.


The object function can be formulated as:

\begin{align}
X' &= G_{\phi}(Mask(X)) \\
\max_{\phi} (\mathcal{L}_G(\phi)) &= D_{MSE}(T(X'),S_{\theta}(X')),
\end{align}

where $X=\{x_{i}\}_{i=1}^{T}$ is the input sequence and $i$
is sequence length.
$Mask(.)$ is a function that randomly masks tokens of the input sequence $X$. In practice, we mask 30\% percent of tokens.
$G_{\phi}(.)$ is the adversarial generator network with parameter $\phi$, $T(.)$ and $S_{\theta}(.)$ are the teacher and student respectively, $D_{MSE}$ is the mean squared error.



The student is trained in two phases. During phase 1, the student model will only learn from the teacher. During phase 2, however, the student model will learn from the ground-truth label.

In phase 1, we anneal the teacher logits inspired by ~\citet{Aref2021}. Note that the student logits are not annealed. The annealing schedule progressively moves from a lower temperature to a temperature of 1. We thus move from a smoother distribution to a sharper softmax distribution. For phase 1, we train to minimize the following losses:

\begin{align}
\
\mathcal{L}_\text{ADV} &=   D_{MSE}(t  \cdot T(X') ,S_{\theta}(X')), \\
\mathcal{L}_\text{KD} &= D_{MSE}(t  \cdot T(X) , S_{\theta}(X)),
\end{align}

where $T$ is the teacher network, $S$ is the student network, $\theta$ is the set of student parameters, $X'$ is the augmented sample and $t$ is the temperature. A $max_T$ hyperparameter will be introduced to calculate $t$. If the epoch number during the training is smaller than $max_T$, then $t=\frac{epoch}{max_T}$\footnote{During the training, epoch is from $1$ to $Epoch_{max}$ that set as hyperparameter.}; otherwise, $t=1$. The total loss in this phase is $\mathcal{L}_\text{ADV} + \mathcal{L}_\text{KD}$.

For phase 2, the student model $S_{\theta}(.)$ will only learn from original data, and we employ cross entropy (CE) loss as our objective function. 
The complete algorithm can be found in  Appendix B.

\begin{figure}[tb]
    \centering
    \includegraphics[width=7cm]{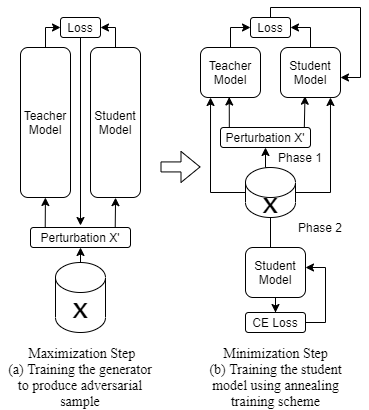}
    \caption{llustration of the maximization and minimization steps of ComKD. For Maximization step, a generator will be trained to generate adversarial samples to maximize the difference between teacher model's and student model's output. For the minimization step, the annealing training scheme will be employed and the student model will learn to match the teacher output on both the original and the perturbed input.}
    \label{fig:ComKD}
\end{figure}

\section{Experiment}

\subsection{Data}
\label{sec:data}

For in-domain test, we evaluate previous KD models as well as our proposed ComKD model on nine datasets of
the General Language Understanding Evaluation
(GLUE)~\cite{wang2019glue} benchmark which includes classification and regression tasks. These datasets
can be broadly divided into 3 families of problems:
1) Single sentence tasks which include linguistic acceptability (CoLA) and sentiment analysis (SST-2).
2) Similarity and paraphrasing tasks which include
paraphrasing (MRPC and QQP) and a regression
task (STS-B).
3) Inference tasks which include Natural Language Inference (MNLI, WNLI, RTE) and Question Answering (QNLI).

For OOD test, we employ HANS~\cite{zhang-etal-2019-paws} and PAWS~\cite{DBLP:journals/corr/abs-1904-01130} as our OOD test set. Models are first trained on MNLI and QQP dataset respectively.

For UAF test, we used GLUE benchmark dev set as our source data, and textfooler~\cite{jin2019bert} as our adversarial algorithm. $\lambda_{down}$ is set to 0.5 and $\lambda_{up}$ is set to 0.99.
Please note that the textfooler can only be applied to classification tasks, so we do not include results on STS-B. We also exclude WNLI because the dataset size is too small. 
The statistics of the UAF test sets are shown in Table~\ref{tab:UAF}. It's it notable that the size of UAF test set for BERT-base and RoBERTa-large is different because the number of models that participate in the test is also different.

\subsection{Evaluation metrics}
On GLUE, we follow the setting of the GLUE leaderboard~\cite{wang2019glue}. Specifically, CoLA is evaluated by Matthews correlation coefficient (MCC), STS-B is evaluated by Pearson correlations, MRPC is evaluated by F1 score, and  the rest of the datasets are evaluated by accuracy. UAF on GLUE employs the same metrics. For OOD test, both F1 score and accuracy are used to evaluate QQP and PAWS dataset, and we use accuracy on HANS and MNLI.

\begin{table}[h!]
\resizebox{\columnwidth}{!}{	
    \begin{tabular}{c|c|c|c|c|c|c|c}
    & CoLA & MNLI &  MRPC & QQP & QNLI & RTE & SST-2 \\ \hline
    BERT-base & 1200  & 6000 & 1200  & 6000  & 6000  & 1200  & 1200 \\ 
    RoBERTa-large & 1000  & 5000 & 1000  & 5000  & 5000  & 1000  & 1000 \\
    \end{tabular}
    }
    \caption{Dataset size for the UAF test sets}
    \label{tab:UAF}
\end{table}

\subsection{Experimental Setting}
\label{Model_setting}

We evaluate the different KD methods on two settings. 
In the first setting, the teacher model is BERT-base~\cite{devlin2019bert} and the student model is initialized with the weights of DistilBERT \cite{sanh2020distilbert}, which consists of 6 layers with a hidden dimension of 768 and 8 attention heads. We find two student model initialization strategies in the literature. Methods such as PKD~\cite{sun2019patient} and ALP-KD~\cite{passban2020alpkd} initialize the weights of the student model with a subset of the teacher weights. Other methods such as Annealing-KD~\cite{Aref2021} and MATE-KD~\cite{rashid-etal-2021-mate} initialize the student model with a pre-trained one such as DistilBERT. We also present a version of ALP-KD which is initialized with a pre-trained model. The pre-trained models are taken from the authors release. The teacher and student are 110M and 66M parameters respectively with a vocabulary size of 30,522 extracted the using Byte Pair Encoding (BPE) \citep{sennrich-etal-2016-neural} tokenization method. 


On the second setting, the teacher model is  RoBERTa-large~\cite{liu2019roberta} and the student is initialized with the weights of DistillRoBERTa \cite{sanh2020distilbert}. RoBERTa-large consists of 24 layers with a hidden dimension of 1024 and 16 attention heads for a total of 355 million parameters. We use the pre-trained model from Huggingface \cite{Wolf2019HuggingFacesTS}. The student consists of 6 layers, 768 hidden dimension, 8 attention heads, with 82 million parameters. Both models have a vocabulary size of 50,265 extracted using BPE. The model hyperparameter and training details are listed in Appendix A.

For the UAF tests we set K to be 1000 for the larger datasets (MNLI, QQP and QNLI) and 200 for the rest.

\begin{table*}[h!]
\center
\small
\renewcommand{\arraystretch}{1.1}
\resizebox{\textwidth}{!}{
    \begin{tabular}{c|c|c|c|c|c|c|c|c|c|c}
& Method   &  CoLA  &   MNLI & MRPC &   QNLI &  QQP     &    RTE      &    SST-2 &    STS-B   &   Avg. score \\ \hline
\multirow{3}*{Baseline}      & BERT-base              &  59.5       &   84.6         &    90.6   &  91.5    &    91.0          &   68.2       & 93.1      &   88.4 & 80.3   \\
                             &DistilBERT              &   51.3       & 82.1           &  90.1    & 89.2     &    88.5          &  59.9        & 91.3      &   86.9 & 77.3\\
                              &Vanilla-KD              &   47.3       & 82.8           &  89.5    & 89.9     & 90.5             &  66.0      & 90.4       &  86.7  &   77.7 \\ \hline
\multirow{2}*{New training}  & PKD                    &  45.7       &   82.1          &  89.3    &  89.3    &  90.7            &  68.2      &   91.5     & 88.6 & 77.9  \\
                             & ALP-KD                 &  47.0        &  81.9         &  89.2     &   89.7   & 90.7             &   68.6     &  91.9      &  88.6   & 78.2   \\
 \multirow{2}*{scheme}       & ALP-KD (DistilBERT)    &  51.8         &   82.9        &   89.9   &  89.9    &     91.2         &   67.5     &   91.4      &  87.3  & 78.7  \\
                             & Annealing-KD           &  55.2         & 83.8          &   90.2   &   89.8   &    91.2          &    67.9    &   92.1     &   87.5   & 79.3 \\ \hline
\multirow{1}*{Data}         &  Tinybert + Aug         &  55.2        &  82.1         &    87.0   &  89.7    &    89.5          &   68.6       &  91.9      &    87.8 & 78.7 \\ 
\multirow{1}*{augmentation}  & MATE-KD                &\textbf{60.4}  &  84.5  &   90.5  &  \textbf{91.2}   & \textbf{91.4}    &    70.0      &  \textbf{92.2}   &   88.5  & 80.5 \\ \hline 
\multirow{1}*{Ours}          & ComKD                  & 59.4& \textbf{84.7} &  \textbf{91.4} &  90.7        & \textbf{91.4}    & \textbf{71.8} & 91.7            & \textbf{89.1}   & \textbf{80.7}     
 
    \end{tabular}}
    \caption{GLUE dev result for different KD models (BERT). The score for the WNLI task is 56.3 for all models and is included in Avg. score. Bold number are the best performance reached by 6-layer models in this table.}
    \label{tab:glue_dev_combine}
\end{table*}

\begin{table*}[h!]
\center
\resizebox{\textwidth}{!}{	
    \begin{tabular}{c|c|c|c|c|c|c}
       & RoBERTa-large &  DistilRoBERTa & Vanilla-KD & Annealing-KD & MATE-KD  & ComKD (Ours) \\ \hline
   CoLA      & 68.1  &    59.7 & 60.9 & 61.7 (54.0) & 65.9 (56.0)  & \textbf{67.4 (58.6)} \\
   RTE       & 86.3  &   69.7  & 71.1 & 73.6 (73.7) &75.0 (75.0)  & \textbf{80.1 (76.6)} \\
   MRPC    &  91.9  &  90.1    & 90.2 & 90.6 (86.0) & 91.9 (\textbf{90.2})  & \textbf{93.0} (89.7) \\
   STS-B    & 92.3  &  88.3  &  88.8 & 89.0 (86.8) & 90.4 (88.0)  & \textbf{91.5 (88.5)} \\
   SST-2    & 96.4 &   89.8   & 92.5  & 93.1 (93.6) & 94.1 (94.9)  & \textbf{95.2 (95.1)} \\
  QNLI      & 94.6 &   89.1   &91.3  & 92.5 (90.8) & \textbf{94.6} (92.1)  &  91.7 (\textbf{92.6}) \\
  QQP      & 91.5  &   90.4   &91.6    & 91.5 (81.2) & 91.5 (81.2)  & \textbf{91.9 (81.4)} \\
  MNLI-m      & 90.2 &   81.9 & 84.1  & 85.3 (84.4) & 85.8 (85.2)  & \textbf{87.2 (85.9)} \\
  WNLI      & 56.3 &   56.3   &  56.3   & 56.3 (65.1) & 56.3 (65.1)  & 56.3 (65.1) \\ \hline
 Avg. score      &85.3 &   79.5   & 80.8  & 81.4 (79.8) & 82.7 (80.8)  & \textbf{83.9 (81.5)} \\
    \end{tabular}
    }
    \caption{Dev set results on GLUE benchmark (RoBERTa). Annealing-KD, MATE-KD and ComKD results in paranthesis is the leaderboard test result. Bold number are the best performance reached by 6-layer models in this table.}
    \label{tab:ComKD_glue}
\end{table*}

\begin{table*}[h!]
\small
\renewcommand{\arraystretch}{1.1}
    \begin{tabular}{c|c|c|c|c|c|c}
    &MNLI-m (Dev) & HANS &  QQP-dev (Acc) & PAWSqqp(ACC) & QQP-dev (F1) & PAWSqqp (F1) \\ \hline
    RoBERTa-large & 90.2 & 78.2 & 91.5 & 43.3 & 88.8 & 48.8   \\
    DistilRoBERTa   & 83.8 & 58.6  &   91.2   &  34.8  & 88.2   &  44.1 \\
    MATE-KD           & 86.3 & 63.6 & \textbf{92.0} & \textbf{38.3} & \textbf{89.2} & \textbf{46.4}\\
    Annealing-KD           & 84.5 & 61.2 & 91.6 & 35.8 & 88.7 & 44.6 \\
    ComKD (Ours)         &  \textbf{87.2} & \textbf{68.6} & 91.6 & 35.2 & 88.7 & 45.0  \\

    \end{tabular}
    \caption{OOD test result ( Bold numbers are the best performance reached by 6-layer models in this table)}
    \label{tab:OOD_test}
\end{table*}

\section{Evaluation}

\subsection{In-domain test}
On Table~\ref{tab:glue_dev_combine}, we present the result of the KD algorithms on GLUE when the teacher is BERT-base. We present an additional baseline for data augmentation following \citet{rashid-etal-2021-mate} that adds the data augmentation from TinyBERT~\cite{jiao-etal-2020-tinybert} to Vanilla-KD. We observe that all the methods improve on the Vanilla-KD results. On the methods which introduce intermediate layer distillation, ALP-KD performs better than PKD. Moreover, initializing ALP-KD with DistilBERT is better than initializing it with the teacher weights. MATE-KD, which employs adversarial data augmentation, performs the best among baseline methods followed by Annealing-KD which anneals the teacher weights. Our proposal, ComKD, which combines both adversarial data augmentation and annealing training scheme outperforms all these methods. The results of MATE-KD indicate that data augmentation is a successful strategy on all datasets and performs particularly well on the smaller ones such as CoLA, STS-B and RTE. 

We evaluate the best performing baselines, Annealing-KD and MATE-KD, as well as our method on the RoBERTa setting. Here, the teacher is RoBERTa-large and the student is initialized with the weights of DistilRoBERTa. Table~\ref{tab:ComKD_glue} presents the dev set results and the test set results (in paranthesis) on GLUE. We see two interesting trends; First, the results follow the same pattern as the previous setup where ComKD is the best, followed by MATE-KD, Annealing-KD and Vanilla-KD. Second, we see a larger gap between our algorithm and MATE-KD. In contrast to MATE-KD we anneal the teacher logits and this has shown to alleviate the capacity gap problem ~\cite{Aref2021}, i.e. a larger capacity difference between the teacher and the student makes distillation more difficult. When learning from a larger teacher, annealing the logits as well as data augmentation both improve KD.

\begin{table*}[h!]
\center
\resizebox{\textwidth}{!}{
    \begin{tabular}{c|c|c|c|c|c|c}
    
       \multirow{2}*{}   &  \multirow{2}*{BERT-base} &  \multirow{2}*{DistilBERT} & ALP-KD  &  \multirow{2}*{Annealing-KD} &  \multirow{2}*{MATE-KD}  &  \multirow{2}*{ComKD (Ours)} \\ 
       &          &          &       (DistilBERT)          &        &           &           \\
       
       \hline
   CoLA      & 39.2 & 24.6 & 26.8 & 25.8 & 35.5 & \textbf{39.7}\\
  MNLI      &  89.8 & 87.3 & 89.4 & 90.6 & \textbf{91.7} & 90.9 \\
  MRPC    &  91.8 & 91.0 & 89.0 & 92.0 & 91.3 & \textbf{92.1}  \\
  QNLI      & 91.1 & 87.9 & 89.6 & 90.0 & \textbf{90.8} & 90.1 \\
QQP      & 90.3 & 86.1 & 84.1 & 85.6 & 85.5 & \textbf{87.4} \\
RTE       & 74.5 & 72.4 & 77.3 & \textbf{81.2} & 75.9 & 74.7 \\
SST-2    &84.4 & 82.3 & 82.1 & 83.8 & 83.8 & \textbf{84.5} \\ \hline
Avg. score (by dataset)   &80.2 & 75.9 & 77.0 & 78.4 & 79.2 & \textbf{79.9} \\
Avg. score (by sample size) & 86.6 & 83.0 & 83.8 & 84.9 & 85.6 & \textbf{85.9} \\
\end{tabular}
}
    \caption{Unified adversarial framework test for BERT-base teacher. Bold number are the best performance reached by 6-layer models in this table.}
    \label{tab:U_b_textfoole1}
\end{table*}

\begin{table*}[h!]
\center
\resizebox{\textwidth}{!}{
\renewcommand{\arraystretch}{1.1}
    \begin{tabular}{c|c|c|c|c|c}
   & RoBERTa-large & DistilRoBERTa & Annealing-KD & MATE-KD & ComKD (Ours)\\\hline
   CoLA      & 14.7 & 5.0 & 2.4 & \textbf{6.6} & 4.9  \\
  MNLI      & 37.0 & 36.6 & 36.6 & 37.0 & \textbf{37.5}  \\
  MRPC    & 94.9 & 90.7 & 88.7 & \textbf{94.2} & 93.4  \\
  QNLI      & 94.2 & 90.8 & 92.4 & \textbf{92.9} & 92.8 \\
QQP      & 89.3 & 86.2 & 87.9 & 87.2 & \textbf{88.1} \\
RTE       & 77.4 & 69.7 & \textbf{73.4} & 69.2 & 71.5 \\
SST-2    & 87.6 & 81.8 & 81.8 & 82.9 & \textbf{84.0} \\\hline
Avg. score (by dataset)  & 70.7 & 65.8 & 66.2 & 67.1 & \textbf{67.5} \\ 
Avg. score (by sample size) & 72.5 & 69.2 & 70.0 & 70.4 & \textbf{70.8} \\
    \end{tabular}
    }
    \caption{Unified adversarial framework test for RoBERTa-large teacher. Bold number are the best performance reached by 6-layer models in this table.}
    \label{tab:U_Rob_textfoole1}
\end{table*}

\subsection{Out-of-Domain test}

We conduct OOD test for RoBERTa-large teacher setting and the are results shown in Table~\ref{tab:OOD_test}.


Specifically, ComKD performs better than MATE-KD on HANS dataset. MATE-KD get better performance on PAWS. To some extend, data augmentation does help the model perform good on OOD tests. Here, we see that the gap between the teacher and student is much larger on the OOD datasets compared to the in-domain testing. Thus, when evaluating the performance of KD and evaluating the gap between teacher and student, we should consider perturbed datasets in addition to the in-domain testing.

To further compare the ComKD and MATE-KD, we introduce the UAF tests which the evaluation sets are generated by each model itself.



\subsection{Adversarial Attack}

In order to compare the adversarial robustness of each KD method, we conduct UAF tests.

\begin{table*}[!thp]
\center

\resizebox{\textwidth}{!}{	
    \begin{tabular}{llcccccccc}
    Index & Sample & Label & RoBERTa-large & DistilRoBERTa  & Annealing-KD & MATE-KD  & ComKD (Ours) \\ \hline
\multirow{2}*{1} &  \textbf{P}:  Yes , you 've done very well , young man & \multirow{2}{*}{C} & \multirow{2}{*}{C} & \multirow{2}{*}{C} & \multirow{2}{*}{C} & \multirow{2}{*}{C} & \multirow{2}{*}{C} \\
  & \textbf{H}: No , you have not done very adequately  &                   &                   &                   &                   &                   &   \\ \hline

\multirow{5}*{2} &  \textbf{P}: All of the islands are now officially  & \multirow{5}{*}{E} & \multirow{5}{*}{N} & \multirow{5}{*}{N} & \multirow{5}{*}{N} & \multirow{5}{*}{N} & \multirow{5}{*}{E} \\
  & and proudly part of France , not colonies  &                   &                   &                   &                   &                   &   \\
  & as they were for some three centuries   &                   &                   &                   &                   &                   &   \\
  
  & \textbf{H}: The island agreed to join France  &                   &                   &                   &                   &                   &   \\
  
  & instead of being colony  &                   &                   &                   &                   &                   &   \\  
  \hline            
  
  \multirow{3}*{3} &  \textbf{P}: I guess history repeats itself , Jane & \multirow{3}{*}{E} & \multirow{3}{*}{N} & \multirow{3}{*}{N} & \multirow{3}{*}{N} & \multirow{3}{*}{E} & \multirow{3}{*}{E} \\
  & \textbf{H}: I truely think the past situation shown  &                   &                   &                   &                   &                   &   \\
  &  history repeats itself  &                   &                   &                   &                   &                   &   \\\hline

 \multirow{2}*{4} &  \textbf{P}:  Case study evaluations & \multirow{2}{*}{E} & \multirow{2}{*}{E} & \multirow{2}{*}{N} & \multirow{2}{*}{N} & \multirow{2}{*}{N} & \multirow{2}{*}{N} \\
  & \textbf{H}: Independent cases studies assessments  &                   &                   &                   &                   &                   &   \\ \hline

  \multirow{3}*{5} &  \textbf{P}: Pretty good newspaper uh & \multirow{3}{*}{E} & \multirow{3}{*}{N} & \multirow{3}{*}{N} & \multirow{3}{*}{N} & \multirow{3}{*}{N} & \multirow{3}{*}{N} \\
  & \textbf{H}: I thinks this is a good newspaper ,   &                   &                   &                   &                   &                   &   \\
  &  and the comics section is my favourite  &                   &                   &                   &                   &                   &   \\\hline    
  
  \hline
    \end{tabular}
    }
    \caption{Some error sample from RoBERTa-large teacher setting UAF test (MNLI). C is contradiction, N is neutral, E is entailment.}
    \label{tab:analysis}
\end{table*}

\begin{table*}[h!]
\center
\resizebox{\textwidth}{!}{	
    \begin{tabular}{c|c|c|c|c|c}
             & RoBERTa large &  DistilRoBERTa  & Annealing-KD & MATE-KD  & ComKD (Ours) \\ \hline
 Add end mark      &  36.4 & 36.0  &  36.3 & 36.4 &  36.7\\
 Remove end mark      & 37.1 & 36.7 & 36.5 & 37.1 & 37.4 \\
    \end{tabular}
    
    }
    \caption{Model performance on UAF (MNLI) after remove or add end mark. (RoBERTa-large)}
    \label{tab:U_end_mark}
\end{table*}


As introduced in Section~\ref{sec:uniontextfooler} and Section~\ref{sec:data}, we use GLUE datasets as source data and textfooler as the adversarial algorithm.
The textfooler algorithm, for a given trained model and dataset, first computes an importance score of the tokens in a sentence. A token is more important if removing it has a greater impact on the model output. Then, it replaces the important tokens with its closest synonyms. In our setting, textfooler will replace at most 15\% of the tokens in a sequence with their synonyms.

Different from OOD tests which is pre-defined, the evaluation set for UAF test is generated by the tested models themselves.
A robust model is expected to handle both adversarial samples that are generated by itself and adversarial samples generated by other models. 
It is notable that the results on BERT setting cannot be compared with the results on RoBERTa setting, because the test sets are different. 

To fairly compare the model's performance, we also show two kind of average scores. The first is average by dataset and this is similar to how GLUE evaluates by averaging the performance on all datasets. The second one is average by sample size where we do a weighted average and weigh the result on each dataset by its size. Thus, larger datasets receive a greater weight.

Tables~\ref{tab:U_b_textfoole1} and ~\ref{tab:U_Rob_textfoole1} present the UAF results for the BERT-base teacher setting and the RoBERTa-large teacher setting respectively. 
We observe that ComKD outperforms other 6-layer methods on average for both settings and achieves a higher score on four out of seven datasets on the BERT setting and three out of seven on the RoBERTa setting. Similar to the OOD results, we observe that the gap between the teacher and student is larger on the UAF test compared to the in-domain test. On the BERT setting ComKD and MATE-KD achieved a higher score than the teacher. 

The performance of all the KD algorithms is consistent with the trend on the in-domain testing. ALP-KD performance is again lower than the other techniques. Overall, our experiments conclude that structural approaches for fine-tuning are not as effective as data-augmentation and progressive learning.

\subsection{Error Analysis}

\begin{table*}[]
\begin{tabular}{l|lll|lll|lll}
\multirow{2}{*}{UAF} & \multicolumn{3}{l|}{Contradiction} & \multicolumn{3}{l|}{Entailment} & \multicolumn{3}{l}{Neutral} \\ \cline{2-10} 
                     & R         & P         & F1         & R        & P        & F1        & R       & P       & F1      \\ \hline
ALP-KD               &0.926	&0.894	&0.907&	0.829	&\textbf{0.965}	&0.892&	0.923&	0.839	&0.879        \\
Annealing-KD         & \textbf{0.946}   &0.893  &0.919  &0.858  &0.960  &0.894  &0.917  &0.875  &0.895    \\
BERT-base            &  0.937  &0.882  &0.908  &0.844  &0.949  &0.894  &0.907  &0.864  &0.885       \\
ComKD (Ours)               &  0.940  &0.903  &0.921  &0.856  &0.961  &0.905  &\textbf{0.929}  &0.869  &0.897       \\
DistilBERT           &  0.924  &0.859  &0.890  &0.819  &0.923  &0.868  &0.881  &0.847  &0.864       \\
MATE-KD           &    \textbf{0.946}  &\textbf{0.918}  &\textbf{0.932}  & \textbf{0.878}  &0.948  &\textbf{0.911}  &0.923  &\textbf{0.885}  &\textbf{0.904}        \\
\end{tabular}
\caption{Models detailed performance on UAF (MNLI) test.}
\label{UAF_STAT}
\end{table*}
\begin{table*}[]
\begin{tabular}{l|lll|lll|lll}
\multirow{2}{*}{In-domain} & \multicolumn{3}{l|}{Contradiction} & \multicolumn{3}{l|}{Entailment} & \multicolumn{3}{l}{Neutral} \\ \cline{2-10} 
                     & R         & P         & F1         & R        & P        & F1        & R       & P       & F1      \\ \hline
ALP-KD               &  0.847  &0.850  &0.848  &0.812  &0.892  &0.850  &0.831  &0.753  &0.790        \\
Annealing-KD         & 0.857  &0.847  &0.852  &0.839  &0.885  &0.861  &0.818  &0.782  &0.799    \\
BERT-base            &  \textbf{0.867}  &0.858  &\textbf{0.862}  &0.840  &0.895  &0.867  &0.833  &0.788  &0.810      \\
ComKD (Ours)               &  0.859  &0.861  &0.860  &0.842  &\textbf{0.899}  &\textbf{0.870}  &\textbf{0.840}  &0.783  &\textbf{0.811}       \\
DistilBERT           & 0.837  &0.824  &0.831  &0.824  &0.865  &0.844  &0.795  &0.767  &0.781       \\
MATE-KD           &   0.858  &\textbf{0.864}  &0.861  &\textbf{0.855}  &0.876  &0.866  &0.819  &\textbf{0.793}  &0.806        \\
\end{tabular}
\caption{Models detailed performance on In-domain (MNLI) test.}
\label{Indomain_STAT}
\end{table*}

In this section, we analyze the error of UAF (MNLI) test.

Table~\ref{tab:analysis} shows some of the UAF samples generated by RoBERTa based models on MNLI. 
For the sample 2, only ComKD can predict correctly. Even though this sample has overlap between the  premise and hypothesis, these models don't predict entailment directly, which indicates the prediction decisions don't only rely on the word overlap. The semantics of words are also important.

For sample 3, both MKD and ComKD can predict correctly, and other models, however, can not. In this sample, there is a length mismatch of premise and hypothesis, as a result, it is harder to predict. For sample 4, Only RoBERTa-large can predict correctly. To get the correct prediction in this sample, the models need to understand that ``evaluations'' has the same meaning here as ``assessments''.
Sample 5 is a sample that none of models predict correctly. Again, there is a length mismatch of premise and hypothesis. 


We also investigate the influence of punctuation on the RoBERTa-large teacher setting. As shown in table~\ref{tab:U_end_mark}, we make two variants of the UAF (MNLI) dataset. For the first setting, we add ``.'' for all the samples that don't have end mark. For the second setting, all the samples' end mark will be removed. According to the table, the end marks do influence the performance of models. Again ComKD perform better than other models in both settings. We also list some samples to show the influence of punctuation in Appendix C.

\normalem

\subsection{Further Discussion}

To find out how KD methods work differently, we looked at the UAF test result (Shown in Figure~\ref{UAF_STAT}) of MNLI dataset, and further analysed the contradiction, entailment and neutral classes.
We can see that \emph{data augmentation based} KD methods (ComKD and MateKD) have higher precision on Contradiction label samples, which means that these model can not be easily confused by negation words since the recall is close for most of KD methods. We see a higher precision in entailment class for ALP-KD, Annealing-KD and ComKD. We also found that KD models perform better than finetune students on each label's f1 score. In summary, \emph{data augmentation based} KD tend to classify Contradiction labels, on the other hand, \emph{better training scheme} KD models prefer to classify Entailment labels. We also see the same trend on In-domain test result (Shown in Figure~\ref{Indomain_STAT}). In both UAF and In-domain results, \emph{data augmentation based} KD methods outperform \emph{better training scheme} KD methods, which indicates that the student trained with data augmentation can achieve a better robustness compared with new training scheme strategy.

\subsection{Conclusion}

In this work, we conduct in-domain, OOD  and UAF test to investigate the robustness of current KD methods. Results show that the KD models' are more robust than fine-tuned student models but less robust than teacher model.
In general, the robustness ranking of each KD methods is consistent with GLUE benchmark average score. 
Specifically, the student trained with data augmentation can achieve a better robustness compared with new training scheme strategy.
Moreover, we also verify that the two strategies of KD methods can be combined together to get a more robust KD model. Our newly proposed ComKD not only outperforms all of the KD methods and achieves SOTA results on the GLUE benchmark, but can also achieve better robustness according to the OOD and the UAF tests.

\section*{Acknowledgements}
We thank MindSpore~\footnote{https://www.mindspore.cn/}, a new deep learning computing framework, for partial support of this
work.





\bibliography{emnlp2021}
\bibliographystyle{acl_natbib}

\clearpage

\appendix

\section{Model training details}
\label{ComKD_details}
For all of the baseline models (Annealing-KD, MATE-KD, ALP-KD, PKD, Vanilla-KD) mentioned, we strictly follow the hyperparameters that are introduced in the original paper. For the DistilBERT student ALP-KD, we only change the initialization of student. The training manner and hyperparemeter tuning is in consistence with original ALP-KD. For the hyperparameters of ComKD that are listed in table~\ref{tab:hyperparameters}, we manually tuned these based on MateKD. Adam optimizer will be applied to train the ComKD.
The generator model that employed in ComKD is following setting of the MATE-KD.
Specifically, for BERT-base teacher setting, a 4-layer Bert-mini model is used.
For RoBERT-large teacher setting, a distilroberta-base model is used.
We trained all models using a single NVIDIA V100 GPU. All experiments were run using the PyTorch~\footnote{https://pytorch.org/} framework.

\begin{table}[h!]
\center
\small
\renewcommand{\arraystretch}{1.0}
    \begin{tabular}{c|c|c|c|c|c}
    & batchsize & lr & ep1 & ep2 & T \\ \hline
   CoLA      & 32 & 7e-6 & 100 &  10 & 10 \\
  MNLI      & 32 & 2e-5 & 30 &  10 & 10 \\
  MRPC    & 32 & 7e-6 & 200 &  10 & 10     \\
  QNLI      & 32 & 2e-5 & 100 &  10 & 10 \\
  QQP      & 32 & 2e-5& 30 &  10 & 10 \\
  RTE       & 32 & 6e-6 & 200 &  10 & 10 \\
  SST-2    & 32 &  1e-5 & 100 &  10 & 10  \\
  STS-B  & 32 &  2e-5 & 100 &  10 & 10\\
    \end{tabular}
    \caption{Hyperparameters for ComKD. ep1 and ep2 is corresponding to the training epochs of phrase 1 and phrase 2. T is max temperature}
    \label{tab:hyperparameters}
\end{table}

\section{Combined-KD Detailed Algorithm }
\label{Algorithm}
In this section, we list Combined-KD details in algorithm~\ref{alg:distillation}.

\begin{algorithm}[tb]
\scriptsize
\tiny

\SetKwInput{Parameter}{Parameter}
 \SetKwInput{pretrain}{pre-trained Student}
 \SetKwInput{teacher}{Finetuned Teacher}
 \SetKwInput{Gen}{Generator model}
 \SetKwInput{data}{dataset}
 \SetKwInput{init}{initialize}
 \SetKwFunction{MSE}{MSE}
 \SetKwFunction{KLD}{KL-divergence}
  \SetKwFunction{CE}{CE}
 \teacher{$T(\cdot)$}
 \pretrain{$S(\cdot; \theta)$}
 \Gen{$G(\cdot; \phi)$}
 \data{$D$}
 \tcc{ Generator training steps, every S steps, max temperature, learning rate}
 \BlankLine
 \Parameter{$S_{g}$, $S$, $max_T$, $\eta$}
$step \leftarrow 0$

\tcc{ learning temperature }
$temp \leftarrow 1$
\BlankLine
\For{$batch \leftarrow D$}
{
    $X, Y \leftarrow batch$ \;
    \BlankLine
        
    $step \leftarrow step \quad \mathrm{mod} \quad S$
    
    \textcolor{blue}{\# Adversarial Step} \;

    $X^m \leftarrow X=[x_1, ..., x_n]$ \;  
    
    $p\sim \text{unif}(0,1), {\text{Mask}(x_i \in X, p_i)}$ \;
    
    \tcc{ predict logit only for the masked tokens }
    $X_\text{logits} \leftarrow G(X^m \phi)$ \;
    
    $X' \leftarrow \text{Gumbel-Softmax}(X_\text{logits})$  \;
    
    \eIf{$step < S_{g}$}{
    $\mathcal{L}_G \leftarrow \MSE(T(X'),S(X';\theta))$
     \tcc{ update generator parameters}
     $\phi \leftarrow \phi - \eta \dfrac{\partial \mathcal{L}_G}{\partial \phi}$ \;
     
  }{
  \textcolor{red}{\# Knowledge Distillation} \;
  
  \eIf{Phase = 1}{
   
  $T_{X'} \leftarrow \frac{temp}{max_T}T(X') $\;
  \BlankLine
  $\mathcal{L}_\text{ADV} \leftarrow   \MSE( T_{X'} ,S(X';\theta))$ \;
  \BlankLine
  $T_{X} \leftarrow \frac{temp}{max_T}T(X) $ \;
  \BlankLine
    $\mathcal{L}_\text{KD} \leftarrow  \MSE( T_{X} , S(X;\theta))$ \;
    \BlankLine
    $\mathcal{L} \leftarrow  \frac{1}{2}\mathcal{L}_\text{ADV} + \frac{1}{2}\mathcal{L}_\text{KD}$ \;
    \BlankLine
    $\theta \leftarrow \theta - \eta \dfrac{\partial \mathcal{L}}{\partial \theta}$ \;
    
    \If{$temp \ne max_{T}$}{
        $temp \leftarrow temp + 1$ \;
}

    }  {
  $\mathcal{L} \leftarrow \CE(S(X;\theta), Y)$  \;
      \BlankLine
  $\mathcal{L} \leftarrow \CE(S_{\theta}(X), Y)$ \;
  \tcc{ update student parameters}
   
  $\theta \leftarrow \theta - \eta \dfrac{\partial \mathcal{L}}{\partial \theta}$ \;
    
  }



    
    \BlankLine
  decay $\eta$ \;
}
}
\caption{Combined Knowledge Distillation }
\label{alg:distillation}
\end{algorithm}

\begin{table*}[h!]
\center
\small
 \renewcommand\tabcolsep{1.0pt}
\renewcommand{\arraystretch}{1.0}
\resizebox{\textwidth}{!}{
    \begin{tabular}{llcccccccc}
    Index & Sample & Label & BERT & DistilBERT  & ALP-KD*  & AKD* & MKD*  & ComKD \\ \hline

\multirow{2}*{1} &  \textbf{P}: I just stopped where I was  & \multirow{2}{*}{E} & \multirow{2}{*}{E} & \multirow{2}{*}{N} & \multirow{2}{*}{N} & \multirow{2}{*}{N} & \multirow{2}{*}{N} & \multirow{2}{*}{N} \\
  &  \textbf{H}: He felt very sick      &                   &                   &                   &                   &                   &                   &   \\\hline
 \multirow{1}*{1 } &  \textbf{P}: I just stopped where I was .  & \multirow{2}{*}{E} & \multirow{2}{*}{E} & \multirow{2}{*}{N} & \multirow{2}{*}{N} & \multirow{2}{*}{\textbf{E}} & \multirow{2}{*}{N} & \multirow{2}{*}{N} \\
 (Add end mark) &  \textbf{H}: He felt very sick .      &                   &                   &                   &                   &                   &                   &   \\\hline
   \multirow{3}*{2} &  \textbf{P}:  the census of 1931 served as an alarm signal for the & \multirow{3}{*}{C} & \multirow{3}{*}{C} & \multirow{3}{*}{N} & \multirow{3}{*}{N} & \multirow{3}{*}{N} & \multirow{3}{*}{C} & \multirow{3}{*}{C} \\
     &   malay national consciousness     &                   &                   &                   &                   &                   &                   &   \\
   
  &  \textbf{H}: there was n't any censuses in malaysia prior to 1940      &                   &                   &                   &                   &                   &                   &   \\ \hline
  
     \multirow{2}*{2 } &  \textbf{P}:  the census of 1931 served as an alarm signal for the & \multirow{3}{*}{C} & \multirow{3}{*}{C} & \multirow{3}{*}{N} & \multirow{3}{*}{N} & \multirow{3}{*}{\textbf{C}} & \multirow{3}{*}{C} & \multirow{3}{*}{C} \\
     &   malay national consciousness .    &                   &                   &                   &                   &                   &                   &   \\
   
 (Add end mark) &  \textbf{H}: there was n't any censuses in malaysia prior to 1940 .     &                   &                   &                   &                   &                   &                   &   \\ \hline

   \multirow{2}*{3} &  \textbf{P}:   oh , what a fool i feel !  & \multirow{2}{*}{C} & \multirow{2}{*}{E} & \multirow{2}{*}{N} & \multirow{2}{*}{N} & \multirow{2}{*}{N} & \multirow{2}{*}{C} & \multirow{2}{*}{C} \\
  &  \textbf{H}: I am beyond pride     &                   &                   &                   &                   &                   &                   &   \\\hline
     \multirow{1}*{3} &  \textbf{P}:   oh , what a fool i feel ! & \multirow{2}{*}{C} & \multirow{2}{*}{E} & \multirow{2}{*}{\textbf{E}} & \multirow{2}{*}{N} & \multirow{2}{*}{N} & \multirow{2}{*}{C} & \multirow{2}{*}{C} \\
 (Add end mark) &  \textbf{H}: I am beyond pride .    &                   &                   &                   &                   &                   &                   &   \\\hline

     \multirow{2}*{4} &  \textbf{P}:  No , don't answer  & \multirow{2}{*}{E} & \multirow{2}{*}{E} & \multirow{2}{*}{C} & \multirow{2}{*}{E} & \multirow{2}{*}{E} & \multirow{2}{*}{E} & \multirow{2}{*}{E} \\
  &  \textbf{H}:  Don't say a word .     &                   &                   &                   &                   &                   &                   &   \\\hline
  
       \multirow{1}*{4} &  \textbf{P}:  No , don't answer . & \multirow{2}{*}{E} & \multirow{2}{*}{E} & \multirow{2}{*}{\textbf{E}} & \multirow{2}{*}{E} & \multirow{2}{*}{E} & \multirow{2}{*}{E} & \multirow{2}{*}{E} \\
 (Add end mark) &  \textbf{H}:  Don't say a word .     &                   &                   &                   &                   &                   &                   &   \\\hline
  
       \multirow{2}*{5} &  \textbf{P}: how long has he been in his present position  & \multirow{2}{*}{E} & \multirow{2}{*}{E} & \multirow{2}{*}{N} & \multirow{2}{*}{E} & \multirow{2}{*}{E} & \multirow{2}{*}{E} & \multirow{2}{*}{E} \\
  &  \textbf{H}: what length of time has he held the current position ?     &                   &                   &                   &                   &                   &                   &   \\\hline
  
         \multirow{1}*{5} &  \textbf{P}: how long has he been in his present position  & \multirow{2}{*}{E} & \multirow{2}{*}{E} & \multirow{2}{*}{\textbf{E}} & \multirow{2}{*}{E} & \multirow{2}{*}{E} & \multirow{2}{*}{E} & \multirow{2}{*}{E} \\
   (remove end mark) &  \textbf{H}: what length of time has he held the current position     &                   &                   &                   &                   &                   &                   &   \\\hline

  \hline
    \end{tabular}}
    \caption{Details of prediction label change of add/remove end mark.  ALP-KD* is ALP-KD (DistilBERT), AKD* is Annealing-KD and MKD* is MATE-KD.}
    \label{tab:Detail_punctuation}
\end{table*}

\section{Discussion of the influence of punctuation }
\label{punctuation}
Some samples of prediction label change of add/remove end mark are shown on Table~\ref{tab:Detail_punctuation}. Most models will not change the prediction after we remove or add a end mark except for Annealing-KD and DistilBERT.

Interestingly, the Annealing-KD can handle the sample 1 and sample 2 correctly after we add the end mark. DistilBERT will also give correct answer for sample 4 and sample 5. These phenomenons indicate that punctuation will give the models a hint to correctly do a classification, and the models make use of it.



\end{document}